%% file: main.tex
\documentclass[letterpaper, 10 pt, conference]{ieeeconf} 	
\IEEEoverridecommandlockouts	
\usepackage[utf8]{inputenc} 
\usepackage[T1]{fontenc}    
\usepackage{hyperref}       
\usepackage{url}            
\usepackage{booktabs}       
\usepackage{amsfonts}       
\usepackage{nicefrac}       
\usepackage{microtype}      
\usepackage{hyperref}

\usepackage{import}
\usepackage{amsmath}

\usepackage{amsthm}

\usepackage{bm}
\usepackage{color}
\usepackage{soul}

\newtheorem{remark}{Remark}

\usepackage{graphicx}
\usepackage{subcaption}
\usepackage{array}

\newcolumntype{C}[1]{>{\centering\let\newline\\\arraybackslash\hspace{0pt}}m{#1}}

\DeclareMathOperator*{\argmin}{arg\,min}

\begin{document}

\import{sections/}{0-Title}

\maketitle
\thispagestyle{empty}
\pagestyle{empty}

\import{sections/}{0-Abstract}

\import{sections/}{1-Introduction}
\import{sections/}{2-RelatedWork}

\import{sections/}{3-Background}

\import{sections/}{5-Methods}
\import{sections/}{6-Experiments}
\import{sections/}{7-Conclusions}

\medskip
\small

\bibliography{bibliography}

\end{document}

%% file: sections/0-Title.tex


\title{Simultaneous Action Recognition and Human Whole-Body Motion and Dynamics Prediction from Wearable Sensors}

\author{Kourosh Darvish$^{1,*}$, Serena Ivaldi$^{2}$, and Daniele Pucci$^{1}$
\thanks{$^{*}$ Corresponding author.}
\thanks{
    $^{1}$ Artificial and Mechanical Intelligence, Center for Robotics and Intelligent Systems, Istituto Italiano di Tecnologia (IIT), Genoa, Italy.
   {\tt\small name.surname@iit.it}
   \newline
    $^{2}$ Inria, Loria, Universit\'e de Lorraine, CNRS, Nancy, France,
    {\tt\small serena.ivaldi@inria.fr}
    }
\thanks{This work has received funding from the European Union's Horizon 2020 research and innovation programmes under grant agreement No 731540 (An.Dy) and No 869855 (SoftManBot), and Italian National Institute for Insurance against Accidents (INAIL) ergoCub project.}
}

%% file: sections/0-Abstract.tex
\begin{abstract}
This paper presents a novel approach to solve simultaneously the problems of human activity recognition and whole-body motion and dynamics prediction for real-time applications.
Starting from the dynamics of human motion and motor system theory, the notion of mixture of experts from deep learning 
has been extended to address this problem. In the proposed approach, experts are modelled as a sequence-to-sequence recurrent neural networks (RNN) architecture. 
Experiments show the results of 66-DoF real-world human motion prediction and action recognition during different tasks like walking and rotating. The code associated with this paper is available at: 
\url{github.com/ami-iit/paper_darvish_2022_humanoids_action-kindyn-predicition}


\end{abstract}

%% file: sections/1-Introduction.tex
\section{Introduction}
\label{sec:introduction}

This paper addresses the problem of simultaneous human whole-body motion prediction and action recognition from wearable sensors.
Given an unfinished set of observed human motion, the prediction should fundamentally respond to two questions for a predefined time horizon in the future: what the human subject will do next in the short-term at the symbolic level, hence a classification problem; how the human subject will do that, i.e., motion prediction as a regression problem.

Prediction of human motion and actions enables many opportunities in various domains of robotics and biomechanics.
When humans collaborate to perform joint actions, they predict each others' actions to coordinate their own decisions and motion \cite{knoblich2011psychological, sebanz2009prediction}.
Similarly, for a successful joint human-robot collaboration, both robots and humans should predict each other actions, allowing them to plan and adapt in advance.
An anticipatory approach lowers the idle time and leads to a more natural and fluent collaboration \cite{Hawkins2014, dragan2015effects}.
Moreover, prediction results, coupled with predictive control approaches, can boost human safety in collaborative workplaces by avoiding the collision of robots with human coworkers.
In another example of a heavy object lifting scenario in a warehouse, employing the prediction results of the workers’ joint torques or workers' future fatigue enable the robots to initiate collaboration and task sharing, i.e., enhancing ergonomics in workplaces  \cite{van2020predicting, ivaldi2017anticipatory}. In this case, the estimation of dynamic information such as interaction forces with the environment is needed.

Another application of human motion prediction and action recognition includes unilateral robot teleoperation in a remote environment to overcome the communication time delay and limited bandwidth \cite{farajiparvar2020brief}.
In a similar direction, human motion prediction allows for the generation of robot motion references  \cite{Viceconte2021ADHERENT}.
In other domains for exoskeletons and prostheses control, integration of human action and motion prediction with the predictive control approaches can enhance the performance and natural motion profile, therefore resulting in a better user experience and comfort \cite{qiu2020exoskeleton}. 
Last but not least, in autonomous cars, the pedestrian motion prediction can reduce the number of accidents and increase the safety \cite{habibi2018context, pfeiffer2018data}.

As mentioned before, human action and motion prediction can be beneficial to various applications and domains. 
According to functional requirements of the target application, prediction time horizon, the desired accuracy, sensory information, and level of details for prediction may vary. 
To estimate the human and environment interaction forces, some works proposed combining the human dynamics physical constraints with the neural networks using videos \cite{ehsani2020use, li2022estimating}. However, those works only estimate the current interaction forces, and estimation results precision does not satisfy many robotic applications requirements, such as exoskeleton control.

In this paper, we take a first step to bridge the gap between model-based and learning-based approaches to identify the mapping from human dynamical states to future human actions and whole-body motion and dynamics information.
The proposed mapping allows for designing a deep neural network (DNN) architecture to solve the two prediction problems simultaneously.
To do so, we have extended the \textit{mixture of experts} (MoE)  approach such that expert outputs predict human motion and interaction forces, and the gating network classifies human actions.
Each expert is enforced to learn a specific human motion generation policy associated with human action, and the gate outputs predict the human future actions. 
This extension is different from the classic MoE where the user does not have control over the gate outputs. Furthermore,  we allow to predict the future ground reactions forces and torques.
The proposed approach permits solving the problems in real-time for a given time horizon in the future. The code associated with this paper is available at: 
\url{github.com/ami-iit/paper_darvish_2022_humanoids_action-kindyn-predicition}

The paper is organized as follows.
Sec.~\ref{sec:RelatedWork} provides the state of the art and Sec.~\ref{sec:BackgroundExtension} presents the paper background.
Sec.~\ref{sec:methods} defines the problem and presents the proposed extension of MoE.
Experiments and results are discussed in Sec.~\ref{sec:experiments}. Conclusions follow in Sec.~\ref{sec:conclusions}.

%% file: sections/2-RelatedWork.tex
\section{Related Work}
\label{sec:RelatedWork}


The problems of human action recognition and action prediction are often solved similarly as a classification problem.
Different supervised learning approaches have been proposed, including Bayesian networks \cite{zhao2019bayesian}, neural networks (NNs) \cite{ji20123d}, Gaussian mixture models and regression \cite{Darvish2021Hierarchical}, and hidden Markov models \cite{moghaddam2013training}.
Similarly, to predict human motion, different approaches based on neural networks including generative adversarial networks \cite{hernandez2019human}, graph convolutional networks \cite{mao2019learning}, dropout auto-encoder LSTM (long short-term memory) \cite{ghosh2017learning}, adversarial encoder-decoder recurrent networks \cite{gui2018adversarial}, convolutional NNs \cite{li2018convolutional}, recurrent neural networks (RNN) \cite{martinez2017human, chiu2019action}, and recurrent encoder-decoder architecture \cite{fragkiadaki2015recurrent} are proposed.
Another common approach in the literature to address the motion prediction problem is based on inverse reinforcement learning (IRL) methods, for example, in \cite{Berret2011Evidence} for a reaching task, or in \cite{mainprice2016goal} for a shared workspace. 

Recently, \cite{kratzer2020prediction} proposed an approach for short-term and long-term motion prediction using RNN and a gradient-based optimization with hand-crafted cost functions to encode environment constraints. Differently from us, in that work, the target position was given to the human subject and algorithm, i.e., human intention (or high-level action) was known. In another work \cite{widmann2018human}, dynamic movement primitives parameterize human motion, and an extended Kalman filter predicts the place and the time of the handover task.



\begin{figure}
\centering
\includegraphics[width=0.98\columnwidth]{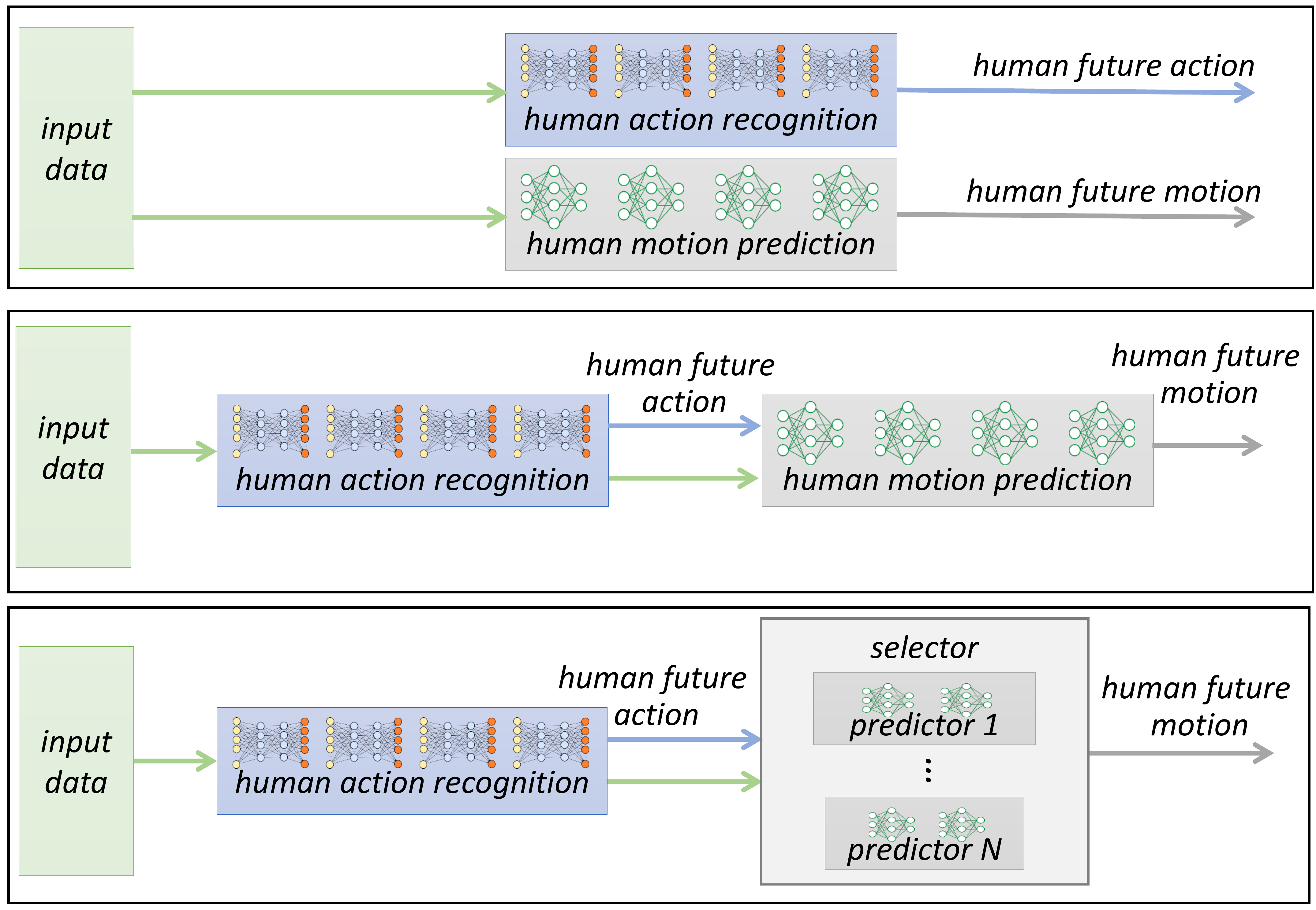}
\caption{Schematic of possible architectures for human action and motion prediction based on supervised learning.}
\label{fig:literature_architectures}
\end{figure}

In the literature, 
many works address solely one among the two problems.
Among those who addressed the two problems together, Fig.~\ref{fig:literature_architectures} indicates different design choices and architectures.
An approach is to solve the two problems separately (i.e., in parallel), as shown in Fig.~\ref{fig:literature_architectures} on top.
However, this design choice neglects the reciprocal correlation of the two problems. This approach introduces the risk that the predicted action and motion do not coincide, i.e., the action $a_i$ is recognized while the predicted motion is related to action $a_j$.
To overcome those problems, one may devise first to predict the human action and provide the result as input (along with other inputs) to the motion prediction problem, as shown in Fig.~\ref{fig:literature_architectures} in the middle.
For example, \cite{luo2019human} probabilistic dynamic movement primitives learn human hand-reaching tasks by inferring first the human intention and then predicting human motion. Yet, in this approach, action prediction results influence the motion prediction, but not reversely.
Extending that, Fig.~\ref{fig:literature_architectures} in the bottom shows an architecture where a single network recognizes human action, and a pool of networks predicts human motion, picked up by a selector  \cite{lasota2017multiple}, similarly to the MoE idea.
This approach may partially untangle generalization problem over different actions; nevertheless, it may introduce a discontinuity problem during the transient phases when human switches from one activity to another. To overcome this, one may consider a weighted summation of motion prediction over different action probabilities.
According to this formulation, the problems of action and motion prediction are not yet mutually interconnected.
To remedy this problem, in \cite{butepage2017deep} an encoder-decoder NN predicts human motion and a part of the same network followed by a fully connected layer classifies human actions. Alternatively, a generative adversarial network can predict human motion whereas a part of the pre-trained discriminator can classify human pose~\cite{barsoum20203d}.

%% file: sections/3-Background.tex
\section{Background}
\label{sec:BackgroundExtension}

To address human action and motion prediction problems, 
this section presents the underlying principles of human motion generation and action from a dynamical system and human motor system perspectives. This study will support the formulation of the two problems with a holistic view, which in turn gives an idea of how to solve them mutually.

\subsection{Human Modeling}
\label{sec:appendix:HumanModling}


Consider a human modeled as a Markov process and is expressed via a multi-body mechanical system with $n$ joints, each with one degree of freedom, connecting $n+1$ links.
Human configuration  is denoted by $\bm{q} = (^{\mathcal{I}}\bm{p}_{\mathcal{B}}, ^{\mathcal{I}}{\bm{R}}_{\mathcal{B}}, \bm{s}) \in \mathbb{R}^3 \times SO(3) \times \mathbb{R}^n$ where $\bm{s}$ is the joint angles, and $^{\mathcal{I}}\bm{p}_{\mathcal{B}}$ and $^{\mathcal{I}}{\bm{R}}_{\mathcal{B}}$ are the floating base position and orientation relative to the inertial frame.
The velocity vector of the model is indicated by $\bm{\nu}=(^{\mathcal{I}}\dot{\bm{p}}_{\mathcal{B}},^{\mathcal{I}}{\bm{\omega}}_{\mathcal{B}}, \dot{\bm{s}}) \in \mathbb{R}^{n+6}$, where its terms are the base linear and rotational (angular) velocity  relative to the inertial frame, and the joint velocity vector.
The velocity of a frame $\mathcal{A}$ attached to a human link, indicated by $ ^{\mathcal{I}}\bm{v}_{\mathcal{A}}= (^{\mathcal{I}}\dot{\bm{p}}_{\mathcal{A}},^{\mathcal{I}}{\bm{\omega}}_{\mathcal{A}}) \in \mathbb{R}^{3} \times \mathbb{R}^{3}$, is computed by its \textit{Jacobian} $\bm{\mathcal{J}}_A(\bm{q}) \in \mathbb{R}^{6 \times (n+6)}$ as
$
^{\mathcal{A}}\bm{v}_{\mathcal{I}}= \bm{\mathcal{J}}_A(\bm{q}) \bm{\nu}
$.
The $n+6$ equations of motion of the human with $n_c$ applied contact wrenches (forces and torques) is ~\cite{sugihara2020survey}:
\begin{equation}
\bm{M}(\bm{q}) \dot{\bm{\nu}} + \bm{C}(\bm{q},\bm{\nu})\bm{\nu} + \bm{g}(\bm{q}) = \bm{B}\bm{\tau} + \sum_{k=1}^{n_c}\bm{\mathcal{J}}_{k}^{T}(\bm{q})\bm{f}^{c}_{k},
\label{eq:dyn}
\end{equation}
with $\bm{M}(\bm{q})$ being the symmetric positive definite inertia matrix, $\bm{C}(\bm{q},\bm{\nu})$ the Coriolis and centrifugal terms, $\bm{g}(\bm{q})$ the vector of gravitational terms, $\bm{B}$ a selector matrix, $\bm{\tau} \in \mathbb{R}^{n}$ the vector of joint torques, and $\bm{f}^{c}_{k} \in \mathbb{R}^{6}$ and $\bm{\mathcal{J}}_{k}$ the vector of the $k$'th contact wrenches and its associated \textit{Jacobian} acting on the human.
\begin{remark}
\label{remark:base-estimation} 
One can show that, by applying a state transformation,  \eqref{eq:dyn} is mapped into an $n+6$ equation where the first 6 equations (centroidal dynamics) depend only on the external wrenches acting on the human, thus being independent from the human internal joint torques \cite{orin2013centroidal}. Furthermore, the last  $n$ equations (free-floating system) of \eqref{eq:dyn} can be expressed only with respect to joint positions and velocities using the rigid contact assumption between the human feet and ground.
\end{remark}

Given \eqref{eq:dyn} and Rem.~\ref{remark:base-estimation}, the human joint dynamics writes:
\begin{equation}
\label{eq:state-space}
\dot{\bm{x}} = \mathcal{F}(\bm{x}, \bm{\tau}, \bm{f}^{c}(t)),
\end{equation}
where $\bm{x}= (\bm{s}, \dot{\bm{s}})  \in \mathbb{R}^{2n}$ denotes the states of the human dynamical system, and $\mathcal{F}$ is a nonlinear function derived from \eqref{eq:dyn} that maps the human states, joint torques, and external forces/torques $\bm{f}^{c} \in \mathbb{R}^{6{n_c}}$ to its rate of change of states.

\subsection{Human Motion Generation}
\label{sec:appendix:HumanMotionGeneration}

According to the literature on biomechanics and motor system and
human dynamics,
we can write down the way a human generates new joint torques as a function of current $\bm{s}(t)$, $\dot{\bm{s}}(t)$, $\ddot{\bm{s}}(t)$, $\dddot{\bm{s}}(t)$ (joint jerks), $\bm{f}^{c}(t) \in \mathbb{R}^{6{n_c}}$ ($n_c$ external forces),  ${\bm{\tau}}(t)$, $\dot{\bm{\tau}}(t), \ddot{\bm{\tau}}(t)$ (the first and second derivative of joint torques resulted from muscle contractions), $\int{\bm{\tau}}^\mathsf{T}(t) \bm{\tau}(t) d(t)$ (joint efforts), $\int{\dot{\bm{s}}}^\mathsf{T}(t) \bm{\tau}(t) d(t)$ (kinetic energy of the joints), and $\bm{r}(t) \in \mathbb{R}^{n_r}$ (other $n_r$ terms associated with the generation of joint torques) \cite{Berret2011Evidence}.
Some of the important terms that we can identify associated with $\bm{r}(t)$ are the human objective or the immediate task, social interaction constraints \cite{adeli2020socially}, the task space constraints such as obstacles, time constraints, and spatial constraints.
In many works in robotics where human motion is predicted, $\bm{r}(t)$ is considered to be known implicitly. It is injected into the problem when a human should act in a structured environment or perform a given task sequence. However, in an unstructured environment or when human subjects are not provided with a description of the tasks to execute, some $\bm{r}(t)$ can be considered as a hidden state in a Markov process and is required to be estimated given input data \cite{russell2002artificial, moghaddam2013training, kelley2008understanding}. Others can be retrieved from the sensory data, such as obstacles in the workspace.

\begin{remark}
\label{remark:optimalility-reaching} 
Biomechanical studies tend to show that humans generate motion to minimize a cost function. This cost function combines mechanical energy expenditure (related to joint torques and velocities) and the motion smoothness (related to minimum jerk) while executing a reaching task \cite{Berret2011Evidence, hogan1984organizing}.
\end{remark}

Following Rem. \ref{remark:optimalility-reaching}, 
the human policy for joint torque generation can be approximated as an optimal control problem with an unknown cost function $\mathcal{J}$ and subject to \eqref{eq:state-space}:
\begin{equation}
\begin{array}{c}
\begin{aligned}
\bm{\tau}^\ast (t)= \underset{\tau(t)}{\argmin}~ \mathcal{J}( \bm{x},{\bm{\tau}}, \ddot{\bm{s}}, \dddot{\bm{s}}, \bm{f}^{c}(t), \dot{\bm{\tau}}, \ddot{\bm{\tau}}, \\
 \cdots,  \int{\bm{\tau}}^\mathsf{T} \bm{\tau} dt, \int{\dot{\bm{s}}}^\mathsf{T} \bm{\tau} dt, \bm{r}(t))
\end{aligned}
\\ \\
s.t. ~~
\dot{\bm{x}} = \mathcal{F}(\bm{x}, \bm{\tau}, \bm{f}^{c}(t))
~,~ \mathcal{C}(.) \leq 0,
\end{array}
\label{eq:optimal-joint-torque}
\end{equation}
where $\mathcal{C}(.)$ is the vector of all inequality constraints.

%% file: sections/5-Methods.tex
\section{Methods}
\label{sec:methods}


\subsection{Problem Statement}
\label{sec:problemStatement}
Following the description of human motion generation and dynamics, here we formalize the problems of human action and motion prediction. 
In this regard, first human dynamics and optimal control problem are discretized.

By discretizing \eqref{eq:state-space} and considering the optimal joint torques obtained from \eqref{eq:optimal-joint-torque}, we can write it as:
\begin{equation}
      {\bm{x}}_{k+1}^\ast = \mathcal{F}(\bm{x}_k, \bm{f}^{c}_k, \bm{\tau}^\ast_k) \Delta{t} +{\bm{x}}_k,
    \label{eq:state-space-discretized}
\end{equation}
where $\Delta{t}$ is the discretization time step.
Moreover, by discretizing \eqref{eq:optimal-joint-torque} and taking advantage of the recursive relationship between the current and previous joint torques, one can compute the optimal joint torques generated at each step by:
\begin{equation}
\begin{aligned}
\bm{\tau}_k^\ast= \mathcal{G}^\ast(\bm{x}_k, \bm{x}_{k-1}, \bm{x}_{k-2}, \dots, \bm{x}_{k-N}, \\ \bm{f}^{c}_k, \dots, \bm{f}^{c}_{k-N}, \bm{r}_k, \dots, \bm{r}_{k-N}).
\end{aligned}
\label{eq:optimal-joint-torque-discretized}
\end{equation}
In this formula, $\mathcal{G}^\ast$ is an unknown and optimal mapping with regard to \eqref{eq:optimal-joint-torque}, $N$ is the number of time steps to look behind in time.
Finally, replacing $ \bm{\tau}^\ast_k$ in \eqref{eq:state-space-discretized} with  \eqref{eq:optimal-joint-torque-discretized}, we can derive the following nonlinear optimal formulation:
\begin{equation}
\begin{aligned}
    {\bm{x}}_{k+1}^\ast = \mathcal{H}^\ast(\bm{x}_k, \bm{x}_{k-1}, \bm{x}_{k-2}, \dots, \bm{x}_{k-N}, \\ \bm{f}^{c}_k, \dots, \bm{f}^{c}_{k-N}, \bm{r}_k, \dots, \bm{r}_{k-N}),
    \end{aligned}
    \label{eq:optimal-state-space-discretized}
\end{equation}
where $\mathcal{H}^\ast$ is an unknown optimal nonlinear function, mapping the input terms to the next state vector $\bm{x}_{k+1}^\ast$.
In this formula, the input terms $\bm{x}_{k-i}$, $\bm{f}^{c}_{k-i}$, and $\bm{r}_{k-i}$ are the states of the system, the vector of external wrenches acting on the human body, and the vector of hidden states at $i$-steps in the past.

By recursively applying \eqref{eq:optimal-state-space-discretized}, we can predict the future states of the human dynamical system for the time horizon $T$, i.e., $\bm{x}_{k+1}^\ast, \bm{x}_{k+2}^\ast, \dots, \bm{x}_{k+T}^\ast$. However, to estimate the future states of the human system in a recursive fashion, there are the following problems that needs to be addressed: ${i})$ the mapping $\mathcal{H}^\ast$ in \eqref{eq:optimal-state-space-discretized} is unknown;
${ii})$ external forces/torques acting on the human in the future $ \bm{f}^{c}_{k+i}$ in \eqref{eq:optimal-state-space-discretized} are not known;
${iii})$ the hidden states  $\bm{r}_{k\pm i}$ in \eqref{eq:optimal-state-space-discretized} are not known, neither in the past nor the future.

\subsection{Guided Mixture of Experts}
\label{sec:GMoE}
To address the challenges derived at the end of Sec. \ref{sec:problemStatement} for human motion prediction, we propose a learning-based approach, i.e., 
the mapping $\mathcal{H}^\ast$ in \eqref{eq:optimal-state-space-discretized} is learned from human demonstrations. As discussed in the literature, approaches based on a single neural network have been proposed to learn the mapping $\mathcal{H}^\ast$. 
However, $\mathcal{H}^\ast$ can be very complex, and yet no approach has resolved this problem effectively.
Starting from \eqref{eq:optimal-state-space-discretized}, here first, we reformulate the action and motion prediction problems in a new form.
Afterward, we adopt the Mixture of Experts (MoE) approach to solve the two problems simultaneously \cite{shazeer2017outrageously, jacobs1991adaptive}.

In order to predict the external wrenches acting on the human in the future $\Tilde{\bm{f}}_{k+i}$, one can come out with two approaches. First, given the predicted states of the human $\hat{\bm{x}}_{k+i}$, we can model the human and the world and perform simulations to predict the external forces acting on the human \cite{koenig2004design}. However, this solution can be time-consuming and it may be cumbersome to model the human and the world for different scenarios.
Another approach is to learn a model of the world for relevant tasks from the human offline demonstrations and try to predict the interaction forces/torques acting on the human \cite{ha2018world}.
For this work, we have decided to go for the learning approach.

In regard to $\bm{r}_{k\pm i}$, when the human subject is not asked to do a given task, the problem becomes even more complex
and depends on many variables. For example, for daily-life activities, to estimate what a human will do and how will do 
them, we should know the hidden internal objective (state) of the human in his mind. Using other sensory modalities like cameras, we may infer the human action, e.g., reaching an object, and human motion and trajectory, e.g., depending on the object's location and obstacles.
However, this is out of the scope of this work, and we are only considering the human dynamical states and interaction forces measured by proprioceptive sensors.
Moreover, depending on the type of $\bm{r}_{k\pm i}$, we can consider $\bm{r}_{k\pm i}$ as the solution of a classification or a regression problem.
In this work, as a simplifying assumption, we only consider human symbolic actions as the hidden state, and will estimate it as a classification problem. In the offline phase, human actions are annotated by experts, while in the online phase, given the input data
human next action is estimated, i.e., $\mathcal{P}(\bm{a}_{k+1}|\bm{x}_k, \dots, \bm{x}_{k-N}, \bm{f}^{c}_k, \dots, \bm{f}^{c}_{k-N})$. Noticeably, in \eqref{eq:optimal-state-space-discretized}, $\bm{r}_k, \dots, \bm{r}_{k-N}$ are compacted and approximated as $\Tilde{\bm{a}}_{k+1}$.
Hence, equation \eqref{eq:optimal-state-space-discretized} can be revised as follows:
\begin{subequations} \label{eq:recognition-prediction-problem}
\begin{align}
  \begin{split}
\Tilde{\bm{a}}_{k+1} &= \mathcal{D}_{1}^\ast(\bm{x}_k, \bm{x}_{k-1}, \bm{x}_{k-2}, \dots, \bm{x}_{k-N}, \\
& ~~~~~~~~~\bm{f}^{c}_k, \dots, \bm{f}^{c}_{k-N}), \label{eq:recognition-prediction-problem-1}
\end{split} \\
\begin{split}
\Tilde{\bm{x}}_{k+1}, \Tilde{\bm{f}^{c}}_{k+1}&= \mathcal{D}_{2}^\ast(\bm{x}_k, \bm{x}_{k-1}, \bm{x}_{k-2}, \dots, \bm{x}_{k-N}, \\
&~~~~~~~~~\bm{f}^{c}_k, \dots, \bm{f}^{c}_{k-N}, \Tilde{\bm{a}}_{k+1}),
\end{split}
\label{eq:recognition-prediction-problem-2}
\end{align}
\end{subequations}
where $\mathcal{D}_{1}^\ast$ and $\mathcal{D}_{2}^\ast$ are two optimal mappings to learn.
As presented, the original complex problem of motion prediction introduced in \eqref{eq:optimal-state-space-discretized} is transformed into action recognition in \eqref{eq:recognition-prediction-problem-1} and motion prediction in \eqref{eq:recognition-prediction-problem-2} problems.
Given these mappings, the problem of motion prediction depends on the problem of action recognition at each inference step.
As described in \eqref{sec:problemStatement} and shown in \eqref{eq:optimal-state-space-discretized}, the problems of action and motion prediction can be solved in a recursive fashion, i.e., repeat the process for the future time horizon $T$.
Consequently, those two mappings are inherently interconnected, and motion prediction results affect the future action recognition results, and at each time step action recognition influence the motion prediction.
Instead of recursive fashion, inference can be performed directly for all the future time horizon $T$ to predict human actions and motion. This way, we expect to enhance the computational time and be suitable for real-time applications, as it computes the outputs for all the time horizon in the future in a tensor form. Nevertheless, as explained before, this approach should meet the interconnection between human action and motion prediction.

\begin{figure}
\centering
\includegraphics[width=0.9\columnwidth]{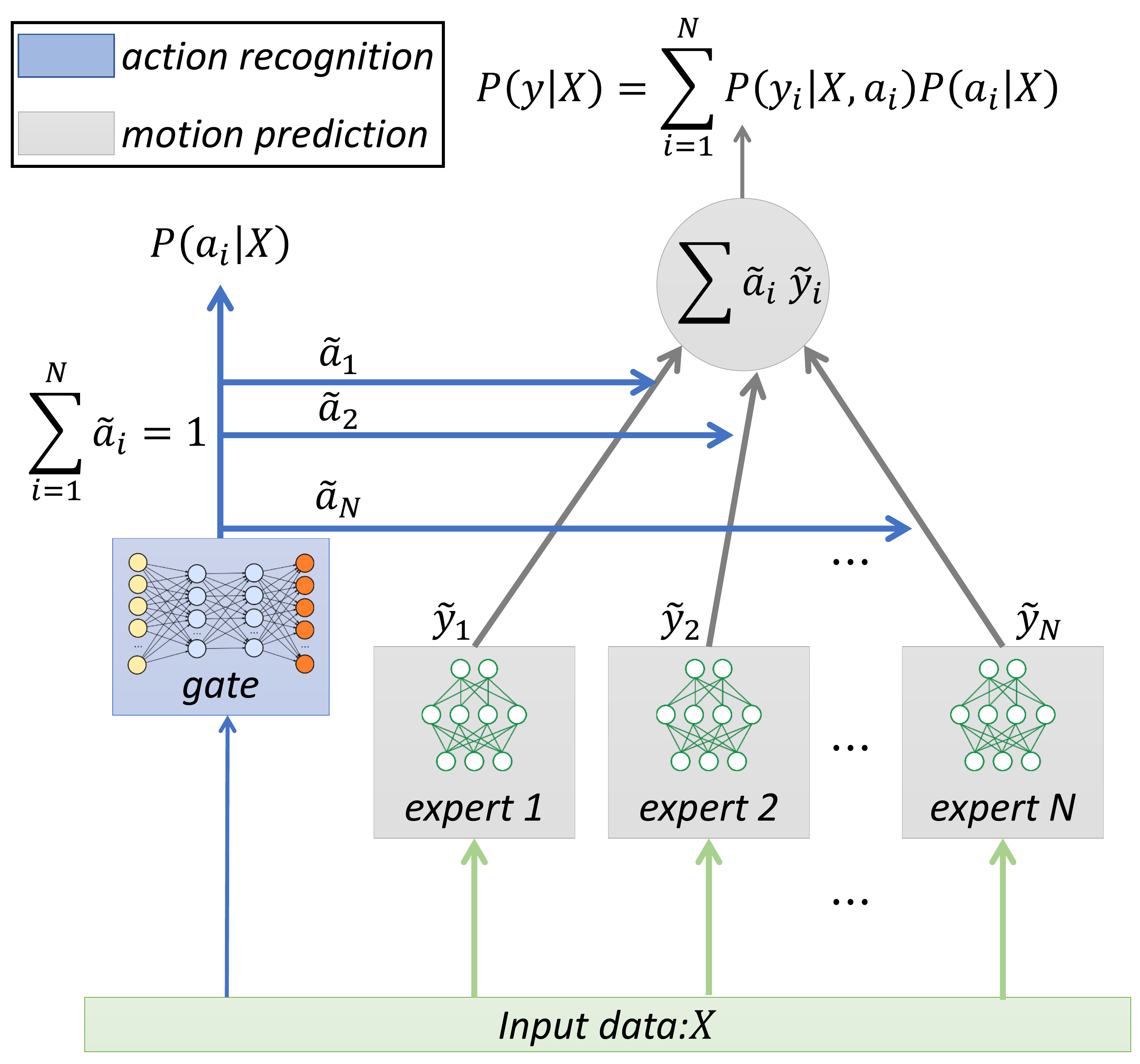}
\caption{Proposed Guided Mixture of Experts (GMoE) for human action and motion prediction. The gate network predicts human action and experts predict human motion.}
\label{fig:MoE_Architecture}
\end{figure}

In order to solve the problem of human action recognition (i.e., learn $\mathcal{D}_{1}^\ast$ in \eqref{eq:recognition-prediction-problem-1}) and motion prediction (i.e., learn $\mathcal{D}_{2}^\ast$ in \eqref{eq:recognition-prediction-problem-2}) jointly, we have elaborated on the idea of MoE as shown in Fig.~\ref{fig:MoE_Architecture}. This proposed architecture is different from both the classical MoE proposed in \cite{jacobs1991adaptive} and the architecture shown in Fig.~\ref{fig:literature_architectures} in the bottom.
In MoE architectures, the gate outputs are not directly controlled, i.e., there is no control on the gate outputs. On the other side in Fig.~\ref{fig:literature_architectures} bottom, the architecture is composed of two sets of NNs, first, the human action recognition is learned, and then the output of the action recognition network and the input data are used to learn human motion prediction (fed to the motion predictors). Hence, it does not consider the inherent and mutual interconnection action and motion prediction as explained in the previous paragraph.
So, as shown in Fig.~\ref{fig:MoE_Architecture}, the two explained shortcomings are addressed with the proposed architecture, \textit{guided mixture of experts} (GMoE).
We consider the outputs of both the gating and expert networks as the two sets of outputs of a single and large MoE network. 
The gate output predicts the human action as a classification problem, while the expert outputs predict human motion as a regression problem.
The gate behavior is guided or controlled via enforcing it to predict human actions and as a result, each expert is trained to learn the motion associated with an action.
As shown in Fig.~\ref{fig:MoE_Architecture}, in the training phase, the gate weights are learned such that they minimize both the error of human action and motion prediction, while the expert weights are learned such that they minimize only the human motion prediction error.
This approach intrinsically allows for smooth transient phases, resolving one of the challenges mentioned in \ref{sec:RelatedWork}. This will be discussed in the experimental results and discussions.

Given the description, the action prediction gate output is $\mathcal{P}(a_i|X)$ where $X$ is the input vector and  $a_i$ is the $i$-th action. $i$-th expert output associated with action $a_i$ can be written as $\mathcal{P}(y_i|X, a_i)$. 
Therefore, the probability distribution of the motion prediction can be written as the marginal probability over the gate outputs as:
\begin{equation}
    \mathcal{P}(y|X) = \sum_{i=1}^{N}\mathcal{P}(y_i|X, a_i)\mathcal{P}(a_i|X),
    \label{eq:experts-summation}
\end{equation}
where $y$ is the motion prediction output vector.



The total loss function $L$ for GMoE can be written as a linear combination of the two output losses $L_1$ (associated with action prediction loss function) and $L_2$ (associated with motion prediction loss function) with the gains $b_1$ and $b_2$ that are set by the user. Here, $L_1$ is set as a categorical cross-entropy loss and $L_2$ is the mean squared error. In other problems, the user may choose different loss functions.
In our case, we define the total loss as $L= b_1L_1 + b_2~L_2$, namely:
\begin{equation}
\begin{array}{l}
\begin{aligned}
    L &=  - \frac{b_1}{2M} \sum_{t=1}^{T} \sum_{j=1}^{M} \sum_{i=1}^{N} a_{i}^{j,t} log(\Tilde{a}_{i}^{j,t}) ~~~~~~~~~~~~~~~~~~~~~~~ \\
    & + \frac{b_2}{2M} \sum_{t=1}^{T} \sum_{j=1}^{M} \|\Tilde{\bm{y}}^{j,t}-\bm{y}^{j,t}\|_2 , 
    s.t.~~ \Tilde{\bm{y}}^{j,t}=\sum_{i=1}^{N} \Tilde{a}_{i}^{j,t} \Tilde{\bm{y}}_{i}^{j,t} ,
\end{aligned}
    \label{eq:GMoE-loss-function}
    \end{array}
\end{equation}
where scalar value $a_i^{j,t}$ and vector $\bm{y}_i^{j,t}$ are human action and motion (e.g., joint values, joint velocities, reaction forces) ground truth
related to the $i$-th action and $j$-th data at the time instance $t$ in the future, and $\Tilde{\cdot}$ indicates estimated values that are stochastically found.
$M$ is the total number of data, and $N$ is the number of experts or modeled actions. When designing the network, $b_1$ and $b_2$ are positive numbers chosen manually as hyperparameters such that both classification (action recognition) and regression (motion prediction) problems converge while training. For this purpose, a suggested approach is first to tune $b_1$ such that the classification problem converges, and later accordingly, set the parameter $b_2$. In this way, we are ensuring that each expert is learning the motion associated with an action. Moreover, $l_1$ and $l_2$ regularization terms can be used to penalize the weight values and avoid overfitting, however they are not reported in the loss function in \eqref{eq:GMoE-loss-function}.
Looking at \eqref{eq:GMoE-loss-function}, during back-propagation while training, we can observe that the gate weights rely on both $L_1$ and $L_2$ losses, while the expert weights only depend on $L_2$.
This shows an important feature of the proposed approach: not only does the human action affect how the human moves, but also the way the human motion affects the recognized action.
Moreover, when the human subject is performing an action $a_i$ (assuming the optimization problem is converged), $a_k$ goes close to zero $\forall k \neq i$, hence the $i$-th expert is enforced to learn the human motion associated with $i$-th action. Finally, in the transient phase when the subject alters from an action to another, the two associated experts try together to reduce the error on the motion prediction output, proportional to the gate outputs.

%% file: sections/6-Experiments.tex
\section{Experiments, Results \& Discussions}
\label{sec:experiments}

\begin{figure}
\centering
\includegraphics[width=1.0\columnwidth]{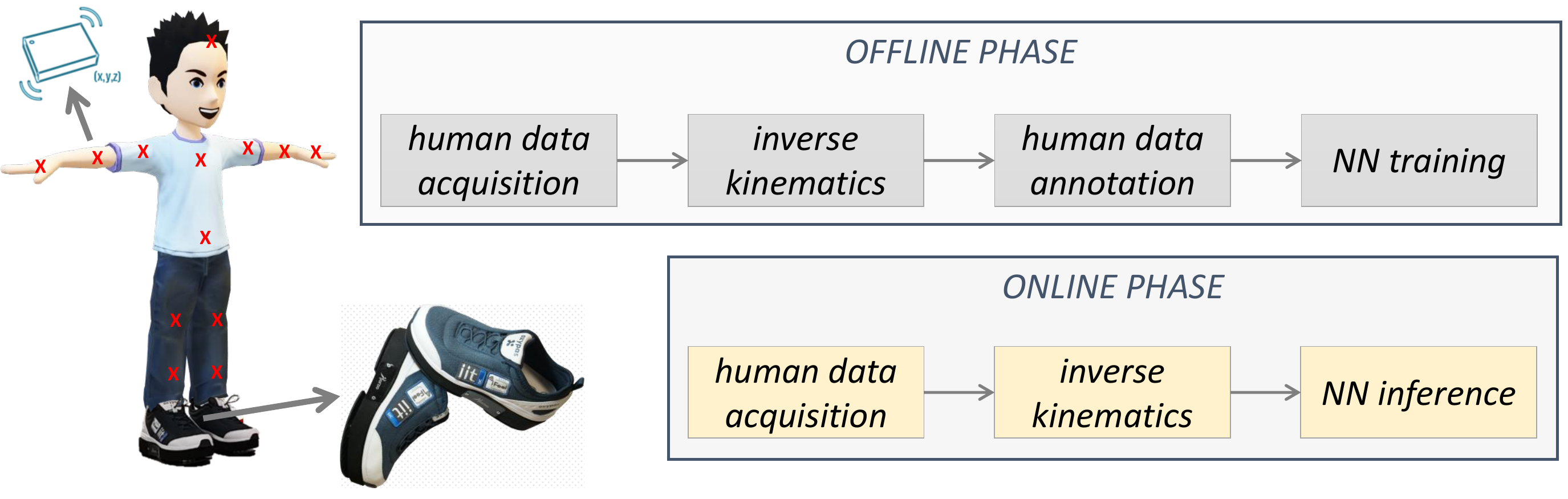}
\caption{Offline and online setup pipelines.}
\label{fig:ExperimentalSetup}
\end{figure}
\subsection{Experimental setup}
The hardware experimental setup and software pipeline are shown in Fig.~\ref{fig:ExperimentalSetup}.
In this setup, human data are collected using Xsens wearable motion capture system \footnote{https://www.xsens.com/} which streams the inertial measurement unit (IMU) sensors data connected to each body link of the subject. The ground reaction forces and torques are measured using iFeel shoes equipped with force/torque sensors \footnote{https://ifeeltech.eu/}.
The data are streamed through a wearable device \footnote{https://github.com/robotology/wearables} using the YARP middleware \cite{metta2006yarp}.
Humans are modeled using a 66 DoFs URDF model, and an inverse kinematic implementation computes the joint values and velocities \cite{rapetti2020model} \footnote{https://github.com/robotology/human-dynamics-estimation}. Data are resampled at 25 Hz.
In the offline phase, for NN training the human whole-body data are visualized and annotated, and are logged to train GMoE NN architecture later.
Instead, in the online phase, the inverse kinematics outputs and shoes data are streamed to the NN inference block in order to estimate future human motion for a given time horizon.
In the online phase, both the ground truth data and predicted ones are visualized.
The programs run on a 64 bit i7 2.8 GHz workstation, equipped with 32 GB RAM, Ubuntu 20.04 LTS, and Intel(R) Iris(R) Xe Graphics.

During experiments, human subject was asked to walk naturally inside a room space, and in total less than $8~mins$ of data have been collected and carefully annotated. The human subject was doing the following actions: \textit{Walking}, \textit{Rotating}, \textit{Stopping}, and other irrelevant actions labeled as \textit{None}.
 In this case, 70\% of data is considered as the training data, 20\% validation data, and the last 10\% as the test dataset.
GMoE architecture is implemented in TensorFlow2 using the functional API. Four similar experts (with one LSTM layer) associated with the number of human actions and one gate network (with two Dense layers) have been considered.
For comparison purposes, an architecture with four LSTM layers for action recognition and motion prediction is implemented as well, similar to Fig.\ref{fig:literature_architectures} on the top.
While training, the Adam optimizer with a decayed learning rate is used. Moreover, to overcome overshooting problems, dropout and batch normalization layers are used in the implemented architecture.
Finally, the inputs to the network are joint values and velocities, and ground reaction forces/torques with $N=5$ past data in \eqref{eq:recognition-prediction-problem}.
Since LSTMs are inherently recursive, we predict the human motion directly (no autoregressive implementation) for the future time horizon of $1~sec$, i.e., $T=25$ steps.

\subsection{Results}
\label{sec:results}

\begin{figure}
\centering
\begin{subfigure}[b]{0.48\columnwidth}
\includegraphics[width=\columnwidth]{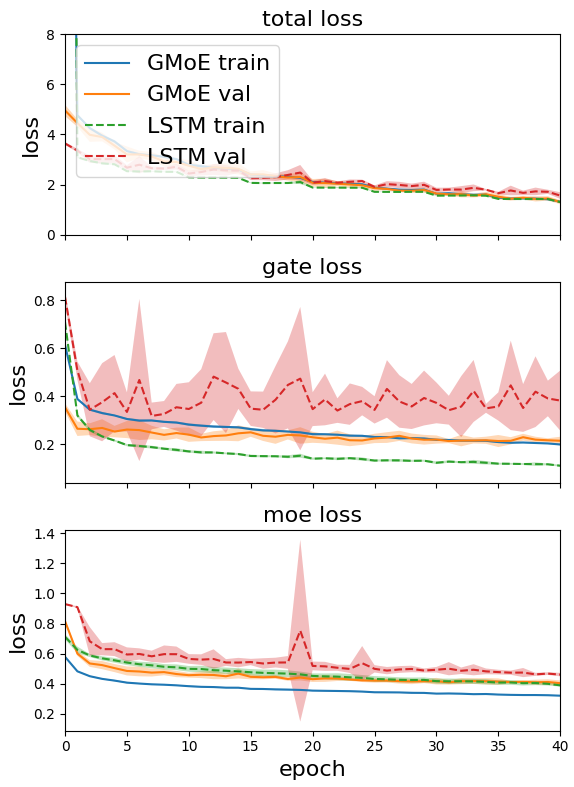}
\caption{}
\label{fig:lossResults}
\end{subfigure}
~
\begin{subfigure}[b]{0.48\columnwidth}
\includegraphics[width=\columnwidth]{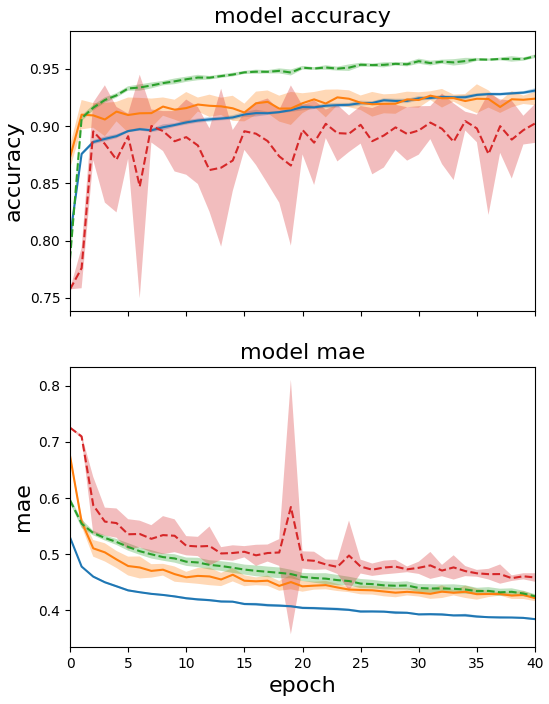}
\caption{}
\label{fig:AccMAEResults}
\end{subfigure}
\caption{Training and validation set results of GMoE and LSTM architectures related to loss functions and metrics for action and motion prediction.
}
\label{fig:Training_Validation}
\end{figure}

\begin{table}[!t]
\caption{Test set mean and standard deviation results of GMoE and LSTM architectures.}
\setlength\tabcolsep{6.0pt} 
\begin{center}
\begin{tabular}{@{}C{1.6cm}C{1.5cm}C{1.5cm}C{1.5cm} @{}}
\toprule
\textit{Architecture} & \textit{total loss}	&  \textit{accuracy} 	& \textit{mae}\\
\hline
\textit{GMoE} & $2.15\pm{0.32}$ & $0.78\pm{0.02}$ &  $0.48\pm{0.02}$\\
 \textit{LSTM} &$2.74\pm{0.42}$ & $0.72\pm{0.05}$ & $0.52\pm{0.02}$\\
\bottomrule
\end{tabular}
\end{center}
\label{tab:GMoE-LSTM-Test-Results}
\end{table}

The mean and standard deviation results of training and validation sets over 10 trials are shown in Fig.~\ref{fig:Training_Validation} for both LSTM and GMoE architectures.
In these experiments, the parameters of \eqref{eq:GMoE-loss-function} are set to $b_1=1.0$ and $b_2=0.2$, and the patience number is set to $5$ while training.
Fig.~\ref{fig:lossResults} on the top shows the total losses related to $L$ in \eqref{eq:GMoE-loss-function}, including $l_1$ and $l_2$ regularization terms as well; in the middle, it shows the action prediction loss related to $L_1$ in \eqref{eq:GMoE-loss-function}, and at the bottom, it shows the loss associated with the motion prediction $L_2$ in \eqref{eq:GMoE-loss-function}.
Fig.~\ref{fig:AccMAEResults} on the top shows the accuracy of action prediction, and at the bottom, it shows the mean absolute error (\textit{mae}) of motion prediction.   
Table \ref{tab:GMoE-LSTM-Test-Results} demonstrates the results of the two architectures on the test set. As shown, even if LSTM architecture has a deeper network with $5.35~millions$  trainable parameters with respect to GMoE with $2.21~millions$ number of trainable parameters, the performance of GMoE surpasses the LSTM architecture.

\begin{figure*}
  \centering
  \includegraphics[width=0.98\textwidth]{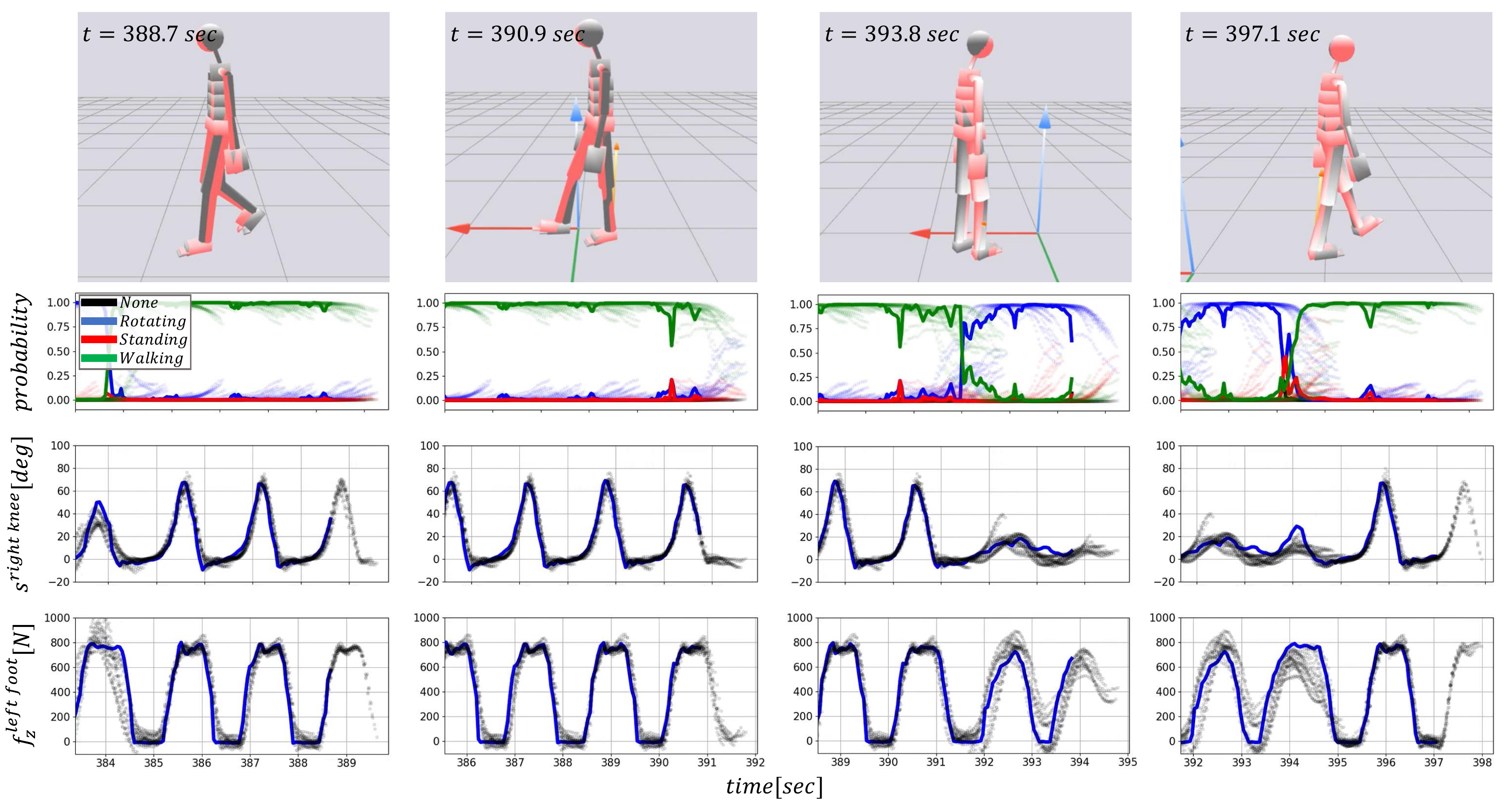}
  \caption{Snapshots of human motion prediction (top), action prediction probabilities (second row), right knee joint angle (in degrees, second row), and left foot ground reaction force (in $N$, bottom); video:
  \href{https://youtu.be/uNs_L2X30xY}{\color{blue}{https://youtu.be/uNs\_L2X30xY}}.}
  \label{fig:results}
\end{figure*}



Fig~\ref{fig:results} shows the results of the human action and motion prediction at different moments.
Online inference takes $30 ms$ on average at each time step running on the specified machine. On the top, it shows the snapshots of the human motion in light gray color and the results of the prediction for $0.2 sec$ in the future in the light red color. Notice that, currently, the future base pose is not estimated, hence the two avatar bases coincide.
In the second row of the figure, black, blue, red, and green colors indicate \textit{none}, \textit{rotating}, \textit{standing}, and \textit{walking} actions.
The results of $T= 1 sec$ of the prediction time horizon are shown with small circles, and probabilities of the next estimated actions are drawn with solid lines.
Finally, figures in the third row and at the bottom demonstrate the results of the prediction of the human right knee joint angle in degrees and the left foot ground reaction force in $z$ direction of the body frame.
In these rows, small circles show the prediction results for the future time horizon at each step, and the solid lines show the current measured values.

In Fig~\ref{fig:results}, at $t=388.7 sec$ (on the left) while the subject is walking, GMoE predicts human will walk for the next $1 sec$ with high probability (close to $1.0$). Hence, human motion prediction predicts the motion associated with the \textit{walking} action for the human for the next $1 sec$.
In the second figure on the left at $t=390.9 sec$, as soon as first data arrives that showing the trace of human starting the \textit{rotating} action, the inference outputs reflect it on the action prediction results, i.e., smoothly the probability of \textit{rotating} action increases (blue color) compared to \textit{walking} action (green color) probability which decreases in the future.
When the human starts to rotate at $t=391.6 sec$, the probability of the human \textit{rotating} action at $t+1.0sec$ is higher than the one at $t$, and reversely for the \textit{walking}.
Later, at $t=393 sec$ human is predicted to rotate for the next $T$ time horizon.
Finally, at $t= 394 sec$ (the fourth column on the right side), the prediction results show a trend from \textit{rotating} to \textit{walking} action for the future time horizon.
For $t \in [392, 393]$, first knee joint angle and feet wrenches is predicted with a \textit{walking} pattern, while later this has been transformed to a \textit{rotating} pattern as the human starts to rotate. This is why in the figure, the predicted joint angle trajectory alter from \textit{walking} trajectory to \textit{rotating} trajectory smoothly.
As denoted by the figure, one of the reasons that the inference results are very sensitive is due to the fact that only the last 5 time steps (i.e., $0.2 sec$) are used to predict the next $1 sec$. Finally, the results of the last row of the figure validate that the proposed architecture predicts accurately the M-shape pattern of human walking stride, which is of paramount importance for biomedical applications.

\subsection{Discussions}
\label{sec:discussions}
In Sec~\ref{sec:BackgroundExtension}, the problem definition is formulated and inspired by the human dynamics and human motor system theory, and encoded motion and interaction forces as shown in Fig.~\ref{fig:results} predict accurately ground truth. However, the proposed solution in the current form does not explicitly take into account human dynamics, i.e., there is no task to constrain the human dynamics, and it cannot ensure the feasibility of the predicted motion. Hence, in future development, we are considering proposing a physics-informed NN to predict the human motion 
\cite{raissi2019physics, li2022estimating, ehsani2020use}.


Connected to the cost function proposed in \eqref{eq:GMoE-loss-function}, however $L_2$ term encourages the associative learning of the experts and discourages the localization of the experts as stated by \cite{jacobs1991adaptive}, the first term in \eqref{eq:GMoE-loss-function} related to $L_1$ encourages the localization of the experts. To further encourage the competitiveness among the experts, one can use other loss functions as $L_2$ in \eqref{eq:GMoE-loss-function}, for example 
$\sum_{i=1}^{N} \Tilde{a}_{i}^{j,t} \| \Tilde{\bm{y}}_{i}^{j,t} - \bm{y}^{j,t}\|_2$ \cite{jacobs1991adaptive}.
In this case, we expect the results of the action prediction do not change considerably while affecting the motion prediction results, especially at transient phases when human action alters.





%% file: sections/7-Conclusions.tex
\section{Conclusions}
\label{sec:conclusions}

In this paper, we proposed a novel approach for simultaneous whole-body human action and motion prediction for the short time horizon in the future. It can effectively predict the human interaction wrenches with the ground. The mixture of experts (MoE) notion has been adopted to solve the two problems together, and the results show the effectiveness of the proposed solution for real-time applications.
In the future, we aim at generalizing the proposed approach over several subjects, and at encoding intraclass human action and motion variations, using a hierarchical version of MoE. Finally, we will consider human dynamics in NN architecture explicitly, to ensure the feasibility of the generated motion and consideration of human constraints.

